# Advances in Artificial Intelligence:
# Are you sure, we are on the right track?

**Emanuel Diamant**
VIDIA-mant, Israel
emanl.245@gmail.com    www.vidia-mant.info

*Abstract:* Over the past decade, AI has made a remarkable progress. It is agreed that this is due to the recently revived Deep Learning technology. Deep Learning enables to process large amounts of data using simplified neuron networks that simulate the way in which the brain works. However, there is a different point of view, which posits that the brain is processing information, not data. This unresolved duality hampered AI progress for years. In this paper, I propose a notion of Integrated information that hopefully will resolve the problem. I consider integrated information as a coupling between two separate entities – physical information (that implies data processing) and semantic information (that provides physical information interpretation). In this regard, intelligence becomes a product of information processing. Extending further this line of thinking, it can be said that information processing does not require more a human brain for its implementation. Indeed, bacteria and amoebas exhibit intelligent behavior without any sign of a brain. That dramatically removes the need for AI systems to emulate the human brain complexity! The paper tries to explore this shift in AI systems design philosophy.

Keywords: Intelligence as data processing, Computational intelligence, Deep learning, Intelligence as information processing, Cognitive intelligence, Brainless intelligence.

**Introduction**

There is now a broad consensus that AI research is making impressive advances in many fields of applications. Reports about dramatic achievements reached in the last years are updated persistently, [1], [2]. The most recent one "2014 in Computing: Breakthroughs in Artificial Intelligence", [2], published by MIT Technology Review, summaries the most important achievements of the past year**.** Some of them are worth to be mentioned:

Facebook's new AI research group reports a major improvement in face-processing software; a technique called deep learning could help Facebook understand its users and their data better.

Researchers at Google have created brain-inspired software that can use complete sentences to accurately describe scenes shown in photos—a significant advance in the field of computer vision.

The search company Baidu, nicknamed "China's Google," also spent big on artificial intelligence. It set up a laboratory in Silicon Valley to expand its existing research into deep learning, and to compete with Google and others for talent.

After IBM's Watson defeat over champions of the sophisticated language game Jeopardy, IBM is now close to make a version of Watson's software help cancer doctors use genomic data to choose personalized treatment plans for patients.

A special part of MIT's review is devoted to a machine learning startup called DeepMind (recently purchased by Google for more than $600 million). DeepMind seeks to build artificial intelligence software that can learn when faced with almost any problem. A computer system developed by DeepMind has learned to play seven Atari video games, six better than previous computer systems and three better than human

experts, with no foreknowledge of the games other than knowing that the goal is to maximize its score. To reach their goals, the DeepMind scientists built on and improved a set of techniques known as Deep learning.

According to Wikipedia's definition, Deep learning is a set of algorithms in machine learning that attempt to model high-level abstractions in data by using model architectures composed of multiple non-linear transformations, [3]. Deep learning software works by filtering data through a hierarchical, multilayered network of simulated neurons that are individually simple but can exhibit complex behavior when linked together.

Almost all mentioned above (in the MIT review) companies base their research on such or another version of Deep learning technique. However, only DeepMind proclaims that their mission is to "solve intelligence", [4]. Such a strong-minded statement looks a bit strange keeping in mind that the term "intelligence" is still not really defined in AI research.

Shane Legg, one of the DeepMind founders, wrote on this issue: "A fundamental problem in strong artificial intelligence is the lack of a clear and precise definition of intelligence itself. This makes it difficult to study the theoretical or empirical aspects of broadly intelligent machines", [5].

In a more earlier publication, "A Collection of Definitions of Intelligence", [6], Legg and Hutter have assembled a list of 70+ different definitions of intelligence proposed by various artificial intelligence researchers. There is no consensus among the items on the list. Such inconsistency and multiplicity of definitions is an unmistakable sign of philosophical immaturity and a lack of a will to keep the needed grade of universality and generalization, [7].

You can agree or disagree with the Ben Goertzel's statement about "philosophical immaturity and a lack of a will" in AI research, [7], but you cannot escape the feeling that without a clear understanding about what is "intelligence" any talks about impressive advancement toward it (DeepMind: our mission is to solve intelligence!, [4]) are simply groundless.

Bearing in mind that AI research is directed towards human intelligence simulation, it seems quite reasonable to inquire into human life sciences for suitable references for such a case. It turns out that human life sciences are coping with the same problem – they too see Intelligence as an undefinable entity. In fact, the polemic has divided the life science community for decades and controversies still rage over its exact definition.

To legitimate my assertion, I would like to draw readers' attention to three most authoritative publications in the field of human intelligence: A "Survey of expert opinion on intelligence" (1987), [8];
A "Report of debates at the meeting of the Board of Scientific Affairs of the American Psychological Association" (1996), [9]; and a review of "Contemporary Theories of Intelligence" (2013), [10].

There is a widely shared opinion that human intelligence cannot be defined as a single trait (Spearman's view (1904) on intelligence as a General Intelligence). Theories of Multiple Intelligences are steadily gaining mainstream attention, [10]. Intelligence becomes an umbrella term that embraces (integrates) a multitude of cognitive capabilities (to sense, to perceive and to interpret the surrounding environment; to recognize, to categorize, to generate plans and to solve problems; to predict future situations, to make decisions and select among alternatives; to learn, to memorize, and so on), which all together produce the effect of intelligence.

Another commonly shared conception is that human cognitive capabilities are all a product of human brain activity, [11]. More precisely, a product of the brain's information processing activity, [12].

If you get an impression that you are near the end of the search and a feel of relief and accomplishment had washed over you, do not lie down and relax – our journey is still not over. Does anybody know "What is information?" No, nobody knows what it is. (Despite of the fact that these days it is the most frequently used and widely adopted word).

The notion of "Information" was first introduced by Claude Shannon in his seminal paper "A Mathematical Theory of Communication" in 1948, [15]. Today, Stanford Encyclopedia of Philosophy offers (side by side with Shannon's definition of information) an extended list of other suggestions being considered: Fisher information, Kolmogorov complexity, Quantum Information, Information as a state of an agent, and Semantic Information (once developed by Bar-Hillel and Carnap), [13]. Again, as it was mentioned earlier, multiplicity of definitions is not a sign of well-being. What makes the difference between "intelligence" and "information" is that in the case of "intelligence" we have forced to cope with multiple definitions of multiple types of intelligence while in the case of "information" we have to cope with multiple definitions of a single entity – certainly an encouraging difference.

I think that the Introduction has already went out of bounds and it will be wise to start with the main part of our discussion.

**What is information?**

For the most of my life, I have earned my living as a computer vision professional busy with Remote sensing and Homeland security projects. The main goal of these endeavors was to understand what is going on in the observable field of view. That is, to reveal the information content of an acquired image. How to do this? – Nobody did not knew then. Nobody knows today. The common practice is to perform low-level image processing hoping in some way or another to reach high-level decisions concerned with image objects detection and recognition.

I tried to approach the problem differently. It took me a long time to do this, but at the year 2005, I have published my first definition of information. I will try to share with you my view on the subject, but for the sake of time and space saving I will provide here only a short excerpt from my once published (and mostly unknown) papers. Interested readers are invited to visit my website (http://www.vidia-mant.info ), where more of such papers can be found and used for further elaboration of the topics relevant to this discourse.

Contrary to the widespread use of Shannon's Information Theory, my research relies on the Kolmogorov's definition of information, [14]. According to Kolmogorov, the definition of information can be expressed as follows: **"Information is a linguistic description of structures observable in a given data set".**

For the purposes of our discussion, digital image is a proper embodiment of what can be seen as a data set. It is a two-dimensional array of a finite number of elements, each of which has a particular location and value. These elements are regarded to as picture elements (also known as pels or pixels). It is taken for granted that an image is not a random collection of these picture elements, but, as a rule, the pixels are naturally grouped into specific assemblies called pixel clusters or structures. Pixels are grouped in these clusters due to the similarity in their physical properties (e.g., pixels' luminosity, color, brightness and as such). For that reason, I have proposed to call these structures **primary or physical data structures**.

In the eyes of an external observer, the primary data structures are further arranged into more larger and complex assemblies, which I propose to call **secondary data structures**. These secondary structures reflect human observer's view on the composition of primary data structures, and therefore they could be called **meaningful or semantic data structures**. While formation of primary data structures is guided by objective (natural, physical) properties of the data, ensuing formation of secondary structures is a subjective process guided by human habits and customs, mutual agreements and conventions between and among members of an observer group.

As it was already said, **Description of structures observable in a data set should be called "Information".** Following the explained above subdivision of the structures discernible in a given image (in a given data set), two types of information must be distinguished – **Physical Information and Semantic Information**. They are both language-based descriptions; however, physical information can be described with a variety of languages (recall that mathematics is also a language), while semantic information can be described only by using human natural language.

I will drop the explanation how physical and semantic information are interrelated and interact among them. Although that is a very important topic, interested readers would have to go to the website and find there the relevant papers, which explain the topic in more details. Here, I will continue with an overview of the primary points that will facilitate our understanding of the issue.

Every information description is a top-down evolving coarse-to-fine hierarchy of descriptions representing various levels of description complexity (various levels of description details). Physical information hierarchy is located at the lowest level of the semantic hierarchy. The process of sensor data interpretation is reified as a process of physical information extraction from the input data, followed by an attempt to associate the physical information at the input with physical information already retained at the lowest level of a semantic hierarchy. If such association is achieved, the input physical information becomes related (via the physical information retained in the system) with a relevant linguistic term, with a word that places the physical information in the context of a phrase, which provides the semantic interpretation of it. In such a way, the input physical information becomes named with an appropriate linguistic label and framed into a suitable linguistic phrase (and further – in a story, a tale, a narrative), which provides the desired meaning for the input physical information. (Again, more details can be found on the website).

**Rethinking intelligence**

As it follows from the above, intelligence has to be understood through cognition, cognition has to be understood through information processing, information processing has to be understood as physical and semantic information interaction and cooperation.

On neither of these steps, data or data processing are not considered as a part or a basis of an ongoing process. Indeed, data features are meaningless in our world perception (and judgment). We understand the meaning of a written word irrelevant to letters' font size or style. We recognize equally well a portrait of a known person on a huge size advertising billboard, on a magazine front page, or on a postage stamp – perceptive information is dimensionless. We grasp the meaning of a scene irrelevant to its illumination. We look on the old black-and-white photos and we do not perceive the lack of colors.
The same is true for voice perception and spoken utterance understanding – we understand what is being said irrelevantly to who is speaking (a man, women, or a child). Irrelevant to the volume levels of the speech (loudly or as a whisper). Blind people read Brail-style writings irrelevant to the size of the touch-code.

Semantic information processing has nothing to do with raw data and its features – raw data features were dissolved in physical information (which is later processed in the semantic information hierarchy). Despite all of this, the new wave of AI innovations relies totally on data processing. The MIT Technology Review, speaking about breakthroughs in AI in the year 2014, enumerates prevailing Deep Learning based developments (which are all data-processing ventures) and then describes the latest IBM's feat – "IBM Chip Processes Data Similar to the Way Your Brain Does"! (Really? Ask Google: "brain is an information processing" – 39900 hits, compare this with: "brain is a data processing" – 6 hit!).

As it follows from the preceding discussion, semantics is not a property of the data. Semantics is a property of a human observer that watches and scrutinizes the data. Semantic information is shared among the observer and other members of his community (and that is the common basis of their intelligence). By the way, this community does not have to embrace the whole mankind. This can be even a very small community of several people or so, which, nevertheless, were lucky to establish a common view on a particular subject and a common understanding of its meaning. Therefore, this particular (privet) knowledge cannot be acquired in any other way. (By Machine Learning, for example, by Deep Learning, or other tricks). **Semantic information should be only shared or granted**! There is no other way to incorporate it as the system's reference knowledge base (used for processing/interpreting physical information at the system's input). Therefore, common attempts to formalize semantics and to derive it from input data are definitely wrong.

The form in which semantic information has to be reified is a string of words, a piece of text, a story, a narrative. (That follows from semantic information definition already given above). If we accept this assumption, it will be reasonable to suppose that semantic information processing means some sort of language texts processing. (For humans it is, obviously, human natural language texts, but for plants or bacteria it will be a different kind of language – every living being possess its own intelligence reified as its ability to process semantic information, which is reified in some pra- or proto-language). What implications follow from the statement "semantic information processing means language text processing"? – I do not know (at least at this stage of my research). As to my knowledge, nobody else knows about this not more than I. (Despite there is a well-known research field of computational linguistics, however, the domain of its studies does not overlap with semantic information processing).

In this regard, it will be fair to mention that IBM's Watson designers and Kurzweil's group at Google are working hard on enabling computers to understand and even to speak in natural language. The goals look similar to my – to reach human level language texts processing. But there is a great and essential difference – both teams pinned all their hopes on suitable deep learning algorithms development.

Adoption of the idea that intelligence is an information processing (duty) naturally raise a question: Is the human brain the only proper means to facilitate this purpose? The answer is: **obviously not**! Bacteria and amoebas exhibit intelligent behavior (intentional external world interaction), which certainly requires information processing, but without any sign of a brain or a nervous system. The same is right for invertebrates, plants, animals, and even mammals. If that is right (and that is right!), why AI systems have to emulate human brain-based intelligence? For me the answer is straightforward: AI systems have to emulate information-processing abilities, not the complexity of a human brain!

In the same vein, it would be right to introduce a new notion for intelligence – Brainless intelligence. And to challenge AI designers with the goal of brainless intelligent devices development.

The time is right to introduce another novelty in the playground of intelligence notions: Computer is a data processing machine; therefore, data-processing-based intelligence has been once dubbed **Computational intelligence**. The advent of information-processing-based approach requires an appropriate new name. In my opinion, **Cognitive intelligence** will be the most suitable option. (Because cognition is a result of information processing and, at the same time, the footing base of intelligence).

It is worth to mention, that IBM and other leading R&D companies are planning to develop and introduce in the near future a computing device of tomorrow – the Cognitive Computer! Which will be capable to perform human like Big data volumes handling. A contradiction (common to all AI projects) is placed from the beginning in the foundations of the enterprise: "Cognitive" is juxtaposed with "Computing". "Cognitive", which implies information processing, and "Computing", which implies data processing! The two are signs of opposite actions. The two are incompatible! (But who cares!)

**Conclusions**

Artificial Intelligence was invented at the Dartmouth College meeting in the summer of 1956. Four brilliant scientists (J. McCarthy, M.L. Minsky, N. Rochester, and C.E. Shannon) have worked out and put into operation the idea of AI research. Despite of the famous fathers, AI research has never been skilful enough to reach its goals. The fathers have failed to assess the complexity of the task and mutual contradictions between its basic constituents. (Recall the story how Minsky proposed to hire a student to solve the problem of vision during the student's summer vocations).

Other examples are plentiful: It was assumed that the best-known manifestation of intelligence is human intelligence; therefore, AI's aim was defined to replicate human intelligence. It was also assumed that the brain is the core and the basis of intelligence. Both assumptions are incorrect – intelligence is an attribute of all living beings, and the existence of a brain is not obligatory for intelligence (bacteria and amoebas exhibit intelligent behavior without any sign of a brain).

The computational paradigm that emerged in the second half of the past century has been generally accepted as the prevalent paradigm of the contemporary science. For that reason, we have "computational intelligence" as well as "computational biology" or "computational linguistics". The brain has become regarded as a computing device, that is, a device aimed at number crunching and data manipulation. Therefore, even today, all AI algorithms are devised to process and to operate data. All breakthrough achievements of the last time are data-processing implements, (Deep Learning is assumed as the most prominent among them).

At the same time, it is generally accepted that the brain is an information-processing appliance. The contradiction between the two definitions can be explained by the peculiarities of scientific development in the past century. It is generally accepted that Intelligence of a living being is expressed in his behavior, that is, in his interaction and communication with the environment. Claude Shannon (one of the AI founders) in his seminal "Mathematical Theory of Communication" [15], has defined that what is being conveyed in a communication process is Information. Shannon was aware that this definition of information is applicable only to data communication, and has nothing to say about the meaning of the conveyed message. Shannon's Information Theory has become widespread and popular in almost all fields of scientific research, despite its shortfalls in message semantics handling. The results of this deficiency are well recognized (over the all AI history) – they have derailed AI research permanently and forever.

This paper is aimed to help the AI researchers to understand the roots of their permanent failures. I have introduced here a new definition of information that will certainly help to avoid in the future the prior mistakes.